# Multiscale style transfer based on a Laplacian pyramid for traditional Chinese painting

**Kunxiao Liu, Guowu Yuan\*, Hongyu Liu and Hao Wu**

School of Information Science and Engineering, Yunnan University, Kunming 650504, China

\* **Correspondence:** Email: gwyuan@ynu.edu.cn.

**Abstract:** Style transfer is adopted to synthesize appealing stylized images that preserve the structure of a content image but carry the pattern of a style image. Many recently proposed style transfer methods use only western oil paintings as style images to achieve image stylization. As a result, unnatural messy artistic effects are produced in stylized images when using these methods to directly transfer the patterns of traditional Chinese paintings, which are composed of plain colors and abstract objects. Moreover, most of them work only at the original image scale and thus ignore multiscale image information during training. In this paper, we present a novel effective multiscale style transfer method based on Laplacian pyramid decomposition and reconstruction, which can transfer unique patterns of Chinese paintings by learning different image features at different scales. In the first stage, the holistic patterns are transferred at low resolution by adopting a Style Transfer Base Network. Then, the details of the content and style are gradually enhanced at higher resolutions by a Detail Enhancement Network with an edge information selection (EIS) module in the second stage. The effectiveness of our method is demonstrated through the generation of appealing high-quality stylization results and a comparison with some state-of-the-art style transfer methods. Datasets and codes are available at https://github.com/toby-katakuri/LP_StyleTransferNet.

---

## 1. Introduction

Artistic painting is a long-established artistic expression technique that comprises lots of diverse information. It not only contains the content information, which reflects what is described but also contains the style information which shows how contents are expressed. For example, scenes and objects are content, and the combination of different colors, strokes and lines represents different styles. Style transfer is a helpful image processing technique in computer vision and its goal is to generate a target image with the content structure of an ordinary content image and the style patterns of an artistic painting. Different from the previous traditional style transfer methods that adopt only low-level image features of a target image to achieve the transfer of style textures [1–3], Gatys et al. [4] proposed a seminal image-optimization style transfer algorithm based on deep convolutional neural network, which uses the correlation of features extracted from a pre-trained visual geometry group (VGG) network [5] to iteratively update a white noise image. Johnson et al. [6], and Ulyanov et al. [7]

propose more efficient model-optimization methods that train a feed-forward neural network with a style image and then generate stylized images with this style in real time. Both of these methods can speed up the optimization procedure of training. Adaptive instance normalization (AdaIN) [8], whitening and coloring transforms (WCT) [9], and style-attentional network (SANet) [10] are also model-optimization style transfer methods, but they can transfer patterns of arbitrary style images after a training process with different feature transforms.

After reviewing these methods, we found that almost all of them adopt only colorful and texture-rich western oil paintings as the style images to generate acceptable stylized images. When using these methods to transfer the patterns of traditional Chinese paintings that are not rich in color and texture, too many disharmonious artistic effects are produced in stylized images. One reason is that these colorful western oil paintings are usually filled with various colors and textures, so the stylization results generated by these methods look acceptable, although some colorful textures may be transferred anywhere. However, when the textures of Chinese paintings, which are composed of simple lines and sober colors, are transferred to the wrong place by these methods, the stylization results look messy and unnatural and lose the spirit of Chinese paintings. Another reason is that the original content and style images are only used with a fixed resolution during the training of these methods. Furthermore, the deep features of images extracted from the neural network only work at the same scale. As a result, trivial details such as local content structures are maintained indiscriminately, and essential elements such as overall object shapes are not highlighted. For example, in some cases, the unnatural style textures are transferred to the blank space of content images and make the stylized images look messy and unaesthetic.

Inspired by the universal process of painting creation [11], which first draws simplified shapes of contents from a holistic view and then adds fine local details gradually, the first stage of our method achieves a rough style transfer at a low resolution, and then the details of stylized images are enhanced gradually at a higher resolution in the second stage. Based on the observations from the Laplacian pyramid translation network (LPTN) [12], some image attributes such as illuminations or colors are exhibited on the low-frequency component, and the content details are more related to the higher-frequency component. Therefore, we propose a novel effective multiscale method based on Laplacian pyramid decomposition and reconstruction for the style transfer task of traditional Chinese paintings. LPTN is an image-to-image translation (I2IT) method that transfers the different daytimes or seasons and adjusts the color or illumination in a target image. Different from LPTN which only handles the same image during the whole training process, the goal of our model is to adopt a content image and a style image to generate a new stylized image.

In the first stage of our model, a Style Transfer Base Network is designed to transfer the global coarse style patterns by adopting the low-frequency component at a low resolution. Different from the translation on the low-frequency component of LPTN that only adopts one image based on some simple convolutional layers to get a translated image, the task of our Style Transfer Base Network is to adopt two different images to generate a stylized image based on the architecture of the style transfer method. In the second stage of our model, a Detail Enhancement Network is employed to enhance the details of contents and patterns by adopting the high-frequency component at a higher resolution. In LPTN, an adversarial loss based on the least squares generative adversarial network (LSGAN) objective [13] and a multi-scale discriminator [14] need to be calculated. However, we discard this loss for achieving faster training speed and still obtain high-quality stylization results.

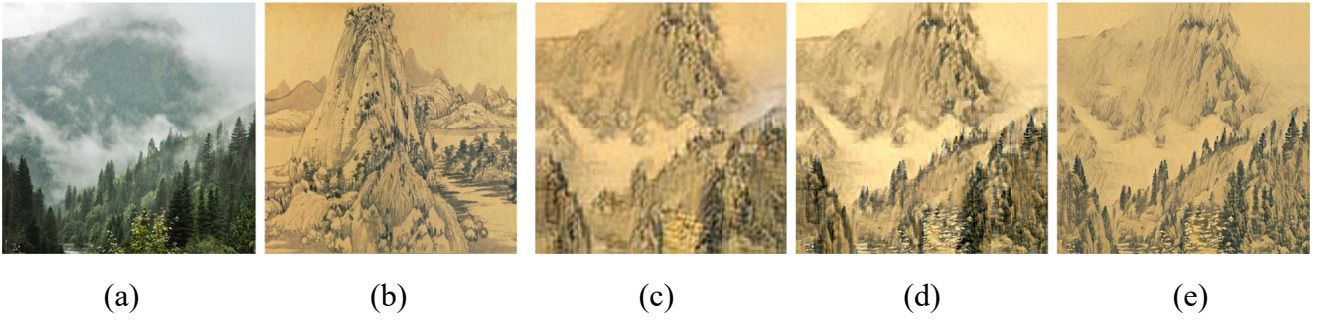

|     |     |     |     |     |
| (a) | (b) | (c) | (d) | (e) |

**Figure 1.** Stylization results generated by our model in different stages. (a) The content image. (b) The style image with Chinese traditional style. (c) The coarse stylized image with a resolution of 128 × 128 is generated by Style Transfer Base Network. (d) The stylized image with a resolution of 256 × 256 is refined in Detail Enhancement Network. (e) The final stylized image with a resolution of 512 × 512 is further enhanced in Detail Enhancement Network.

Figure 1 shows the stylized images with different resolutions synthesized by our model in different stages. In Figure 1(c), a coarse stylized image with a resolution of 128 × 128 is generated by our Style Transfer Base Network in the first stage. The overall color distribution and content structure are transferred roughly in the first stage. Figure 1(d) shows that the texture and color of the stylized image are refined preliminarily at a resolution of 256 × 256 in the second stage. Finally, Figure 1(e) shows that the final high-quality stylized image with a resolution of 512 × 512 is synthesized by our full model in the second stage, where the content details and style patterns are further enhanced.

Our main contributions are as follows:

• We present a novel effective multiscale style transfer method that is designed to transfer the unique style patterns of traditional Chinese painting and then synthesize high-quality stylized images. In the first stage, coarse style patterns such as color distribution and holistic content structure are transferred by using a Style Transfer Base Network at a low resolution. In the second stage, the details of the style patterns and the content objects are gradually enhanced by adopting a Detail Enhancement Network at higher resolutions.

• Laplacian pyramid decomposition and reconstruction are applied in our model for multiscale learning. We use only the low-resolution image to transfer the style in the first stage while avoiding heavy convolutions on feature maps of the high-resolution image in the second stage. In this way, our model avoids the heavy computational cost and can be applied in practical high-resolution applications.

• We propose an EIS module that refines the features of content edge maps to pay more attention to the key semantic information of content structures in the second stage. As a result, essential content details such as the outline of objects can be preserved in the stylized images.

• It is demonstrated through different experiments that our method can transfer the patterns of traditional Chinese paintings and synthesize appealing high-quality stylization results. The style characteristics of Chinese paintings such as plain colors and abstract objects can be transferred accurately. Meanwhile, the structural details of the content images such as the edge of objects can be well maintained.

The following parts of this paper are organized as follows. In Section 2, we review the works related to different style transfer methods based on neural networks. Then, in Section 3, the pipeline of our method, the details of our network, and the loss functions are described. Different experimental results are shown and analyzed in Section 4. Finally, we summarize our work and discuss future research in Section 5.

## 2. Related works

### 2.1. Style transfer

Style transfer is an attractive method of computer vision that is adopted to synthesize the stylized image which preserves the content structure while transferring the style texture. Gatys et al. [4] proposed a pioneering model-optimization style transfer method that adopts a pre-trained VGG network [5] to extract features and update a white noise image iteratively as a stylized image. Recently, Style transfer by relaxed optimal transport and self-similarity (STROTSS) [15] adopted different loss terms to achieve a model-optimization method with multiscale learning, which can generate superior stylization results. However, both of these methods take a lot of time to train and only generate one stylized image after one training. To solve this problem, Johnson et al. [6] proposed a feed-forward method to generate any number of stylized images with the same style after a training process. This model adopts an encoder-decoder architecture. In addition, some approaches [16,17] speed up the style transfer process, and others [18–20] improve visual quality. Sanakoyue et al. [21] trained an autoencoder network with a proposed style-aware loss and a set of style images instead of a style to improve the visual quality of stylized images. Deng et al. [22] proposed a transformer-based framework with two different transformer encoders and a multilayer transformer decoder, which adopt domain-specific sequences of content and style to generate outstanding results based on a content-aware positional encoding (CAPE) scheme. Yang et al. [23] proposed an exemplar-based portrait style transfer method that retains an intrinsic style path to control the style of the original domain and adds an extrinsic style path to model the style of the target extended domain based on a generative adversarial network.

To simultaneously handle multiple styles, a flexible conditional instance normalization (CIN) approach proposed by Dumoulin et al. [24] is applied in the style transfer network to learn multiple styles. Stylized neural painter (SNP) [25] proposed a novel dual-pathway rendering network that can synthesize realistic painting artworks by generating vectorized strokes sequentially to render in a variety of painting styles. In the multi-style method proposed by Ye et al. [26], a convolutional block attention module (CBAM) attention layer and a CIN layer are integrated into the stylization network for preserving the original semantic information. In addition, a YOLACT (you only look at coefficients) network and the Poisson fusion algorithm are used to achieve more natural stylization between different regions.

To generate arbitrary stylized images with arbitrary content and style images, AdaIN [8] adopts adaptive instance normalization to learn multiple styles during training, and WCT [9] applies whitening and coloring transforms to the features of content and style images with a series of pre-trained image restructuring decoders. Inspired by the decomposition of Gram matrices, Wang et al. [27] proposed a deep feature perturbation (DFP) operation to generate diverse results without any extra learning process in WCT-based frameworks. Moreover, SANet [10] achieves the arbitrary style image method by integrating local style patterns into content features based on the self-attention mechanism.

## 2.2. Style transfer for traditional Chinese painting

To transfer the style patterns of traditional Chinese paintings, Lin et al. [28] proposed a deep learning method that uses a generative network with an edge detector to transform sketches into Chinese paintings. Li et al. [29] proposed a neural abstract style transfer method to preserve abstraction for traditional Chinese paintings by adopting different losses based on a novel modified version of the extended Difference-of-Gaussians.

## 2.3. Style transfer based on multiscale learning

For the style transfer framework, working at multiple scales is a useful technique that can capture a wide range of image statistics to improve image quality. A generative adversarial network trained on a single natural image (SinGAN) [30] captures patch-level distribution at different image scales with a pyramid of adversarial networks and then synthesizes high-quality stylized images with a style image. Sheng et al. [31] proposed an Avatar-Net that achieves multiscale holistic style transfer based on an hourglass network with multiple skip connections and a style decorator. Lin et al. [32] proposed a Laplacian pyramid style network (LapStyle) which works at multiple scales and synthesizes stylized images with high visual quality based on a Drafting Network and a Revision Network.

## 3. Proposed method

### 3.1. Framework overview

Unlike most colorful and texture-rich western oil paintings with realism, traditional Chinese paintings are impressionistic and characterized by simple colors and abstract objects. During the process of style transfer, we do not need to preserve too many local content structures while needing only to keep some areas of the stylized images blank and clean. Therefore, we propose to adopt multiscale learning to selectively transfer the style patterns of Chinese paintings while avoiding messy texture generation.

To achieve the training of our model at different scales, we employ the long-standing image processing technique Laplacian pyramid. Specifically, we first use the Laplacian pyramid to decompose an original high-resolution content image into a low-resolution content image as the low-frequency component and a series of high-resolution residual content images as the high-frequency components. Then, a Style Transfer Base Network with two style attentional (SA) modules is designed to transfer style with the low-resolution content image, and a Detail Enhancement Network with an EIS module is designed to refine these residual content images to obtain the set of residual stylized images. Finally, we use the low-frequency component and the set of refined high-frequency components to reconstruct the final high-resolution stylized image based on Laplacian pyramid reconstruction. In this way, the trivial local content structures are discarded while the holistic content structures are maintained at low resolution, and key details of style patterns are enhanced gradually at higher resolution. As a result, our stylization results show the same artistic expression as traditional Chinese paintings.

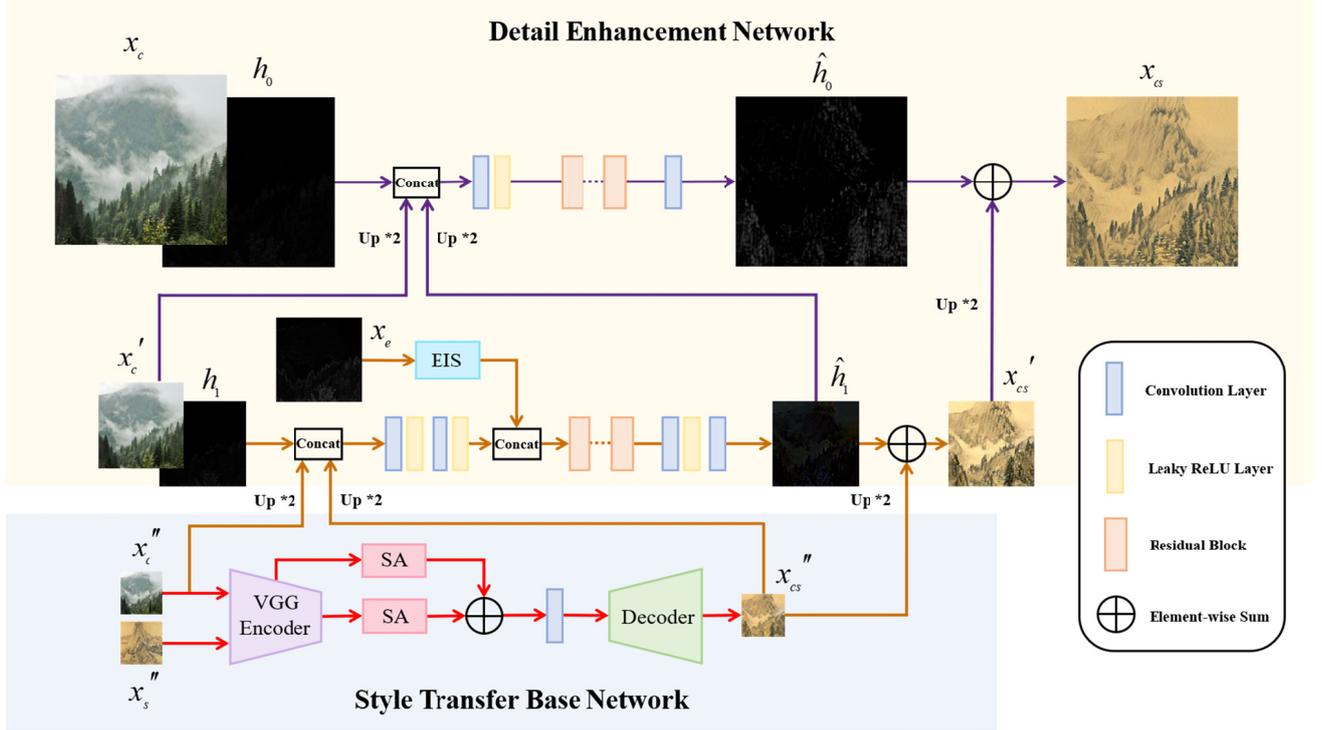

**Figure 2.** Overview of our framework.

Our framework is designed based on a 2-level Laplacian pyramid because of the limitation of computational resources. In Figure 2, given an original content image $x_c \in \mathbb{R}^{3 \times h \times w}$ and an original style image $x_s \in \mathbb{R}^{3 \times h \times w}$, we decompose $x_c$ into a low-frequency component and a set of high-frequency components by adopting Laplacian Pyramid. We downsample $x_c$ to obtain a low-resolution content image $x_c' \in \mathbb{R}^{3 \times \frac{h}{2} \times \frac{w}{2}}$ based on a fixed kernel and further downsample $x_c'$ to obtain a content image $x_c'' \in \mathbb{R}^{3 \times \frac{h}{4} \times \frac{w}{4}}$ with a lower resolution as the low-frequency component. In this way, a set of high-resolution residual content images as the high-frequency components denoted by $H = [h_0, h_1]$ can be obtained, where $h_0 = x_c - Up\left(x_c'\right)$, $h_1 = x_c' - Up\left(x_c''\right)$, and $Up$ is a $2 \times$ upsample operation. In the first stage, our Style Transfer Base Network takes $x_c''$ and $x_s''$ as the input and then synthesizes the coarse low-resolution stylized image $x_{cs}'' \in \mathbb{R}^{3 \times \frac{h}{4} \times \frac{w}{4}}$ as the stylized low-frequency component for reconstruction. In the second stage, our Detail Enhancement Network utilizes two steps to gradually enhance the details. In the first step, the Detail Enhancement Network takes $h_1$, $x_c''$, $x_{cs}''$, and $x_e \in \mathbb{R}^{3 \times \frac{h}{2} \times \frac{w}{2}}$ as the input, where $x_e$ denotes the edge map of $x_c'$. Then, the high-resolution residual stylized image $\hat{h}_1 \in \mathbb{R}^{3 \times \frac{h}{2} \times \frac{w}{2}}$ as the first stylized high-frequency component can be generated. In the second step, our Detail Enhancement Network takes $h_0$, $x_c'$, and $\hat{h}_1$ as the input and then generates the second stylized high-frequency component $\hat{h}_0 \in \mathbb{R}^{3 \times h \times w}$. The descriptions of the components are shown in Table 1. Eventually, the final high-resolution stylized image $x_{cs} \in \mathbb{R}^{3 \times h \times w}$ can be synthesized based on Laplacian pyramid reconstruction by adopting element-wise operations between $x_{cs}''$, $\hat{h}_1$, and $\hat{h}_0$. Specifically, we perform an element-wise operation between $Up(x_{cs}'')$ and $\hat{h}_1$ to obtain $x_{cs}'$ then perform an element-wise operation between $Up(x_{cs}')$ and $\hat{h}_0$ to obtain $x_{cs}$.

**Table 1.** The descriptions of the components.

| Component | Description |
| --- | --- |
| $x_c''$ | Low-frequency component (Low-resolution content image) |
| $x_s''$ | Low-frequency component (Low-resolution style image) |
| $x_{cs}''$ | Stylized low-frequency component (Low-resolution stylized image) |
| $h_0$ | High-frequency component (High-resolution residual content image) |
| $h_1$ | High-frequency component (High-resolution residual content image) |
| $\hat{h}_0$ | Stylized high-frequency component (High-resolution residual stylized image) |
| $\hat{h}_1$ | Stylized high-frequency component (High-resolution residual stylized image) |

In addition, higher-resolution applications can be achieved by increasing the levels of a Laplacian pyramid to handle higher-resolution images. To obtain higher-resolution stylized images, we can stack the network used in the second step in Detail Enhancement Network. Then we repeat the second step for higher-frequency components and generate stylized higher-frequency components to reconstruct.

### 3.2. Style Transfer Base Network

The goal of the Style Transfer Base Network is to transfer coarse style patterns at a low resolution. The low-resolution content image $x_c''$ and the low-resolution style image $x_s''$ are taken as the inputs and the low-resolution stylized image $x_{cs}''$ is generated as the output. Figure 2 shows the architecture of the Style Transfer Base Network, which is designed as an effective encoder-decoder architecture with an encoder, two SA modules, and a decoder. The encoder is a pretrained VGG-19 network, which is fixed during training.

#### 3.2.1. SA module

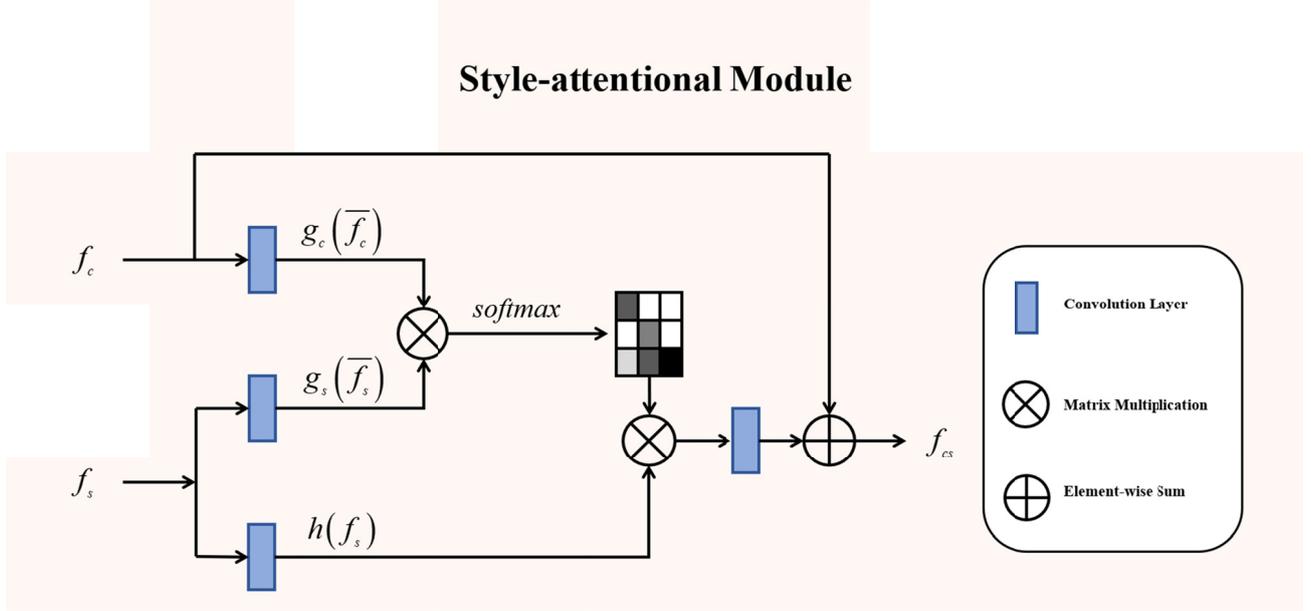

**Figure 3.** Schematic for SA module.

To transfer the rough style patterns in the Style Transfer Base Network, we employ the SA module proposed in [10] to embed the style feature into the content feature. The SA module is designed to preserve the content structure as much as possible while enriching the style patterns based on multilevel feature embeddings. In Figure 3, we take $f_c \in \mathbb{R}^{c_s \times h_s \times w_s}$ and $f_s \in \mathbb{R}^{c_s \times h_s \times w_s}$ as the input of the SA module, where $f_c$ and $f_s$ are the feature maps of low-resolution content image $x_c''$ and style image $x_s''$, which are extracted by a pre-trained VGG network. Next, $f_c$ and $f_s$ are normalized and then transformed into two feature spaces $g_c$ and $g_s$ to calculate the attention between $\overline{f_c^i}$ and $\overline{f_s^j}$ as follows:

$$f_{cs}^i = \sum_{\forall j} \frac{\exp\left(g_c\left(\overline{f_c^i}\right)^T g_s\left(\overline{f_s^j}\right)\right)}{\sum_{\forall j}\exp\left(g_c\left(\overline{f_c^i}\right)^T g_s\left(\overline{f_s^j}\right)\right)} h\left(f_s^j\right) \tag{1}$$

where $g_c\left(\overline{f_c}\right) = W_c\,\overline{f_c}$, $g_s\left(\overline{f_s}\right) = W_s\overline{f_s}$, $h\left(f_s\right) = W_h f_s$, and $\overline{f}$ denotes a mean-variance channel-wise normalized version of $f$. Moreover, $i$ is the index of an output position, and $j$ is the index that enumerates all possible positions. $W_c$, $W_s$, and $W_h$ are the learned weight matrices, which are implemented by convolution layers. Then, $f_{cs}$ obtained by Eq (1) is fed into a convolutional layer to produce a refined result. Finally, we obtain a style-attentional stylized feature $f_{cs} \in \mathbb{R}^{c_s \times h_s \times w_s}$ by performing an element-wise sum operator between this refined result and the original content feature $f_c$, which integrates local style patterns in light of the semantic spatial distribution of the content image.

### 3.2.2. Workflow of Style Transfer Base Network

In Style Transfer Base Network, given $x_c''$ and $x_s''$, the content features $f_c$ and the style

features $f_s$ at ReLU_4_1 and ReLU_5_1 are extracted by the VGG encoder. Next, two SA modules are applied to these features for style feature embedding. We take $f_c^{4-1}$ and $f_s^{4-1}$ from ReLU_4_1 as the inputs of the first SA module and then generate a stylized feature $f_{cs}^{4-1}$. In the same way, the stylized feature $f_{cs}^{5-1}$ can be generated by the second SA module when taking $f_c^{5-1}$ and $f_s^{5-1}$ as the inputs. Then, we add $f_{cs}^{5-1}$ to $Up\left(f_{cs}^{4-1}\right)$ to obtain a multilevel stylized feature and feed it into a convolutional layer to produce a final refined stylized feature $f_{cs} \in \mathbb{R}^{c_s \times h_s \times w_s}$. Finally, the decoder is used to reconstruct $f_{cs}$ to generate the final stylized image $x_{cs}''$, which is designed to be symmetrical to the VGG-19 network.

### 3.3. Detail Enhancement Network

In Detail Enhancement Network, our task is to add the details of style patterns and content structures to stylized images gradually at higher resolution. The high-frequency component $H = [h_0, h_1]$ obtained by Laplacian pyramid decomposition and the content edge map $x_e$ are utilized as the inputs. Then one refined stylized high-frequency component $\hat{h}_1$ is generated in the first step and another refined stylized high-frequency component $\hat{h}_0$ is generated in the second step. Eventually, we take $x_{cs}''$, $\hat{h}_0$, and $\hat{h}_1$ to reconstruct the final high-resolution stylized image $x_{cs}$.

Figure 2 shows the architecture of the Detail Enhancement Network, which is designed to be simple. It includes some simple convolution layers followed by leaky rectified linear unit (ReLU) layers, a series of residual blocks, and an EIS module. The encoding-decoding paradigm with heavy convolutional layers commonly used in some traditional image-to-image translation methods is not used because of the limitations of their applications to high-resolution tasks.

#### 3.3.1. Edge information selection (EIS) module

To preserve more key local content structures such as the outline of objects in stylized images, we adopt the edge map of the content image to contribute semantic content information. An EIS module is designed to select essential semantic features of content images by capturing contextual feature dependencies based on the self-attention mechanism. Inspired by the dual attention network (DANet) proposed in [33], we adopt the channel attention module, which can adaptively integrate local features from a global view in the channel dimensions to model the semantic interdependencies between edge feature maps. In addition, the feature representation of specific semantics can be improved, and interdependent feature maps can be emphasized by exploiting the interdependencies between channel maps. We do not apply the position attention module proposed in DANet in our EIS module due to the limitation of GPU memory.

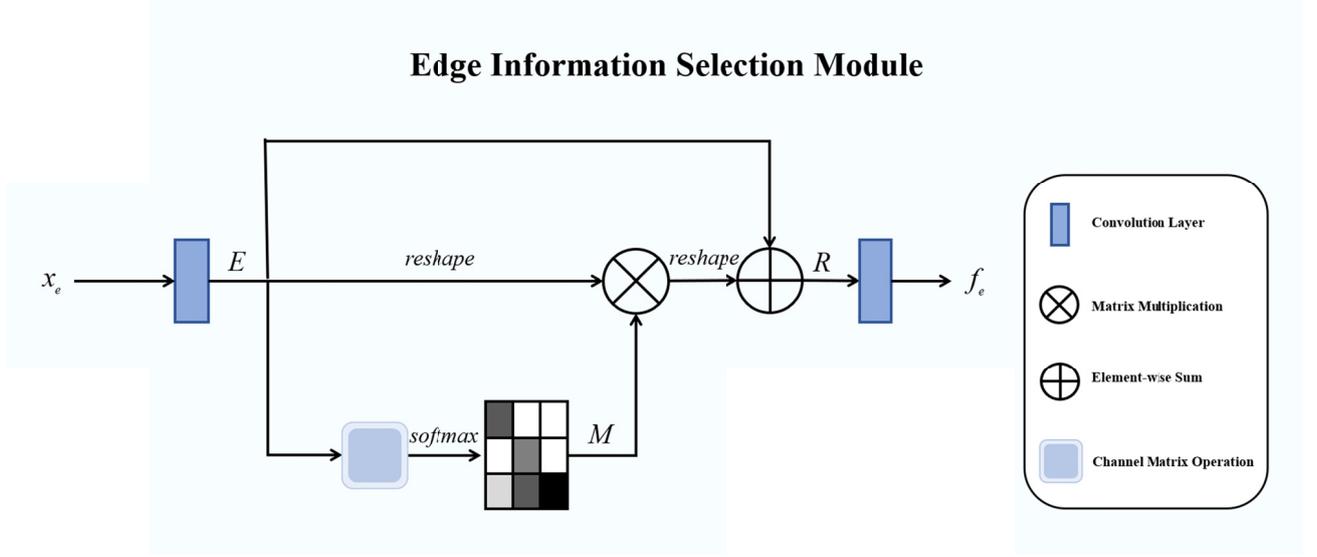

**Figure 4.** Schematic for EIS module.

In Figure 4, we take the edge map of the content image $x_e$ as the input of the EIS module and then feed it into a convolutional layer to produce an edge feature $E \in \mathbb{R}^{c_r \times h_r \times w_r}$. After $E$ is reshaped to $\mathbb{R}^{c_r \times n_r}$, we apply a channel matrix operation that denotes performing a matrix multiplication between $E$ and the transpose of $E$. Then, the channel attention map $M \in \mathbb{R}^{c_r \times c_r}$ is calculated as follows:

$$m_{ji} = \frac{\exp\left(E_i \cdot E_j\right)}{\sum_{i=1}^{c} \exp\left(E_i \cdot E_j\right)} \tag{2}$$

where $m_{ji}$ denotes the $i^{th}$ channel's impact on the $j^{th}$ channel. We perform a matrix multiplication between the transpose of $M$ and $E$ to obtain a result and perform an element-wise sum operator between this result, which is reshaped to $\mathbb{R}^{c_r \times h_r \times w_r}$ and $E$ to produce a refined edge feature $R \in \mathbb{R}^{c_r \times h_r \times w_r}$ as follows:

$$R_j = \beta \sum_{i=1}^{c} \left(m_{ji} E_i\right) + E_j \tag{3}$$

where $\beta$ gradually learns a weight from 0. Eq (3) shows that the refined edge feature of each channel is a weighted sum of the features of all channels and original features. Finally, the refined edge feature $R$ is fed into a convolutional layer to obtain the final edge feature $f_e \in \mathbb{R}^{c_r \times h_r \times w_r}$ as part of the input of the Detail Enhancement Network.

### 3.3.2. Workflow of Detail Enhancement Network

In the first step of Detail Enhancement Network, we concatenate $h_1$, $Up\left(x_c''\right)$, and $Up\left(x_{cs}''\right)$ as the inputs. Meanwhile, we take $x_e$ as the input of the EIS module to produce the edge feature $f_e$. After the second leaky ReLU layer, $f_e$ is concatenated into the network to generate the first refined high-frequency component $\hat{h}_1$. In the second step of the Detail Enhancement Network, we concatenate $Up\left(x_c'\right)$, $h_0$, and $Up\left(\hat{h}_1\right)$ as the input and then generate the second refined high-frequency component $\hat{h}_0$. Eventually, we adopt the set of refined high-frequency stylized components $\hat{H}=\left[\hat{h}_0,\hat{h}_1\right]$ and the low-frequency stylized component $x_{cs}''$ to obtain the final high-resolution stylized image $x_{cs}$ based on Laplacian pyramid reconstruction.

### 3.4. Loss function

Our model includes two training processes for two networks. We first train the Style Transfer Base Network with a single style image in the first stage. Then we train the Detail Enhancement Network with a single style image in the second stage. Style Transfer Base Network is fixed during the training process of Detail Enhancement Network.

#### 3.4.1. Loss function of Style Transfer Base Network

Following SANet [10], we use content, style, and identity losses to optimize our Style Transfer Base Network during training. Given $x_c''$, $x_s''$, and $x_{cs}''$, we use a pretrained VGG-19 encoder to extract their features $F_c^{(t)} \in \mathbb{R}^{c \times h_t \times w_t}$, $F_s^{(t)} \in \mathbb{R}^{c \times h_t \times w_t}$, and $F_{cs}^{(t)} \in \mathbb{R}^{c \times h_t \times w_t}$, where $t$ denotes the features extracted at ReLU_$t$ ($t$ = 1_1, 2_1, 3_1, 4_1, 5_1).

**Style Loss:** For style loss, we use the mean-variance loss as:

$$\mathcal{L}_s = \sum_{\forall t} \| \mu\left(F_{cs}^{(t)}\right) - \mu\left(F_s^{(t)}\right) \|_2 + \| \sigma\left(F_{cs}^{(t)}\right) + \sigma\left(F_s^{(t)}\right) \|_2 \tag{4}$$

where $\mu$ and $\sigma$ denote the mean and covariance of the feature maps, respectively.

**Content Loss:** For content loss, we use the Euclidean distance between $\overline{F_{cs}^{(t)}}$ and $\overline{F_c^{(t)}}$ as follows:

$$\mathcal{L}_c = \sum_{\forall t} \| \overline{F_{cs}^{(t)}} - \overline{F_c^{(t)}} \|_2 \tag{5}$$

where $\overline{F_{cs}^{(t)}}$ and $\overline{F_c^{(t)}}$ denotes a mean-variance channel-wise normalized version of $F^{(t)}$.

**Identity Loss:** To consider both the global statistics and the semantically local mapping between the content features and the style features, we adopt the identity loss function as follows:

$$\mathcal{L}_i = \lambda_{i1} \| x_{cc} - x_c \|_2 + \| x_{ss} - x_s \|_2 + \lambda_{i2} \sum_{\forall t} \left( \| F_{cc}^{(t)} - F_c^{(t)} \|_2 + \| F_{ss}^{(t)} - F_s^{(t)} \|_2 \right) \tag{6}$$

where $x_{cc}$ and $x_{ss}$ denote the output images of our Style Transfer Base Network with two same content images and style images as inputs. $\lambda_{i1}$ and $\lambda_{i2}$ are identity loss weights. In addition, $F_{cc}^{(t)}$ and $F_{ss}^{(t)}$ are the features of $x_{cc}$ and $x_{ss}$ extracted by a pretrained VGG-19 encoder.

The overall optimization objective of the Style Transfer Base Network in the first stage is defined as:

$$\mathcal{L}_{STBNet} = \lambda_c \mathcal{L}_c + \lambda_s \mathcal{L}_s + \mathcal{L}_i \tag{7}$$

where $\lambda_c$ and $\lambda_s$ are weight terms. Specifically, $\mathcal{L}_c$ works on ReLU_4_1 and ReLU_5_1. Then, $\mathcal{L}_s$ and $\mathcal{L}_i$ work on ReLU_1_1, ReLU_2_1, ReLU_3_1, ReLU_4_1, and ReLU_5_1.

### 3.4.2. Loss function of Detail Enhancement Network

To train the Detail Enhancement Network, we adopt content and style losses for optimization. In the same way, we adopt the same pretrained VGG-19 encoder to extract the features of $x_c$, $x_s$, and $x_{cs}$. As a result, we can obtain $F_c^{(t)} \in \mathbb{R}^{c \times h_t \times w_t}$, $F_s^{(t)} \in \mathbb{R}^{c \times h_t \times w_t}$, and $F_{cs}^{(t)} \in \mathbb{R}^{c \times h_t \times w_t}$, where $t$ denotes the features extracted at ReLU_$t$ ($t = 1\_1, 2\_1, 3\_1, 4\_1$).

**Style Loss:** For style loss, we adopt two different style losses. We first adopt the same mean-variance loss used in the Style Transfer Base Network as:

$$l_{mv} = \sum_{\forall t} \| \mu\left(F_{cs}^{(t)}\right) - \mu\left(F_s^{(t)}\right)\|_2 + \| \sigma\left(F_{cs}^{(t)}\right) - \sigma\left(F_s^{(t)}\right)\|_2 \qquad (8)$$

where $\mu$ and $\sigma$ denote the mean and covariance of the feature maps, respectively. Then, we use the relaxed Earth mover's distance (rEMD) loss [15] as the second style loss, which is significant to achieve our final stylization results and improve the visual quality. The rEMD loss between $F_{cs}^{(t)}$ and $F_s^{(t)}$ is defined as:

$$l_r = \sum_{\forall t} \max\left( \frac{1}{h_t w_t} \sum_{i=1}^{h_t w_t} \min_j C_{ij}, \frac{1}{h_t w_t} \sum_{j=1}^{h_t w_t} \min_i C_{ij} \right) \qquad (9)$$

where $C$ is the cost matrix, which can be calculated as the cosine distance between $F_s^{(t)}$ and $F_{cs}^{(t)}$:

$$C_{ij} = D_{cos}\left(F_{s,i}^{(t)}, F_{cs,j}^{(t)}\right) = 1 - \frac{F_{s,i}^{(t)} \cdot F_{cs,j}^{(t)}}{\| F_{s,i}^{(t)} \| \| F_{cs,j}^{(t)} \|} \qquad (10)$$

**Content Loss:** For content loss, we use two different content losses. First, we adopt the same content loss used in the Style Transfer Base Network, which is defined as:

$$l_p = \sum_{\forall t} \| \overline{F_{cs}^{(t)}} - \overline{F_c^{(t)}} \|_2 \qquad (11)$$

Then, we adopt the self-similarity loss [15], which can retain the relative relation in the content image to the stylized image:

$$l_{ss} = \sum_{\forall t} \left( \frac{1}{(h_t w_t)^2} \sum_{i,j} \left| \frac{D_{ij}^c}{\sum_i D_{ij}^c} - \frac{D_{ij}^{cs}}{\sum_j D_{ij}^{cs}} \right| \right) \qquad (12)$$

where $D_{ij}^c$ and $D_{ij}^{cs}$ are self-similarity matrices, which are calculated by pairwise cosine similarity. As shown in Eq (10), $D_{ij}^c = D_{cos}\left(F_{c,i}^{(t)}, F_{c,j}^{(t)}\right)$ and $D_{ij}^{cs} = D_{cos}\left(F_{cs,i}^{(t)}, F_{cs,j}^{(t)}\right)$.

The overall optimization objective of our Detail Enhancement Network in the second stage is defined as:

$$\mathcal{L}_{DENet} = \alpha\left(\lambda_1 l_p + \lambda_2 l_{ss}\right) + \lambda_3 l_{mv} + \lambda_4 l_r \qquad (13)$$

where $\alpha$, $\lambda_1$, $\lambda_2$, $\lambda_3$, and $\lambda_4$ are weight terms. Specifically, $l_p$ and $l_{mv}$ both work on ReLU_1_1, ReLU_2_1, ReLU_3_1, and ReLU_4_1. Then, $l_{ss}$ and $l_r$ both work on ReLU_3_1, and ReLU_4_1.

In addition, by adjusting $\alpha$, the degree of stylization can be controlled.

## 4. Experimental results and analysis

### 4.1. Experimental dataset and implementation details

During both training processes of the Style Transfer Base Network and Detail Enhancement Network, we use the Microsoft common objects in context (COCO) [34] dataset as the set of content images and some famous and representative Chinese traditional paintings as style images. In addition, we select some copyright-free images from Pexels.com as content images to show the experimental results.

In our experiments, we train our Style Transfer Base Network with a set of content images and a single style image in the first stage. Then, the Style Transfer Base Network is fixed, and the Detail Enhancement is optimized to generate the final stylized image in the second stage. In the experiments, we use the content images and the style images with a resolution of 512 × 512. Following SANet [10], we use the Adam optimizer [35] with a learning rate of 1e-4 and a batch size of 5 during the training of the Style Transfer Base Network. Moreover, the weighting parameters $\lambda_{l1}$, $\lambda_{l2}$, $\lambda_c$, and $\lambda_s$ are set to 1, 50, 1, and 3, respectively. During the training of the Detail Enhancement Network, we also use the Adam optimizer with a learning rate of 5e-3, and the batch size is set as 1 because of the limitation of the GPU memory. The loss weight terms, $\alpha$, $\lambda_1$, $\lambda_2$, $\lambda_3$, and $\lambda_4$ are set to 1, 1, 15, 50, and 80, respectively. The experimental environment configuration is shown in Table 2.

**Table 2.** Experimental environment configuration.

| Designation | Information |
| --- | --- |
| Operating system | Windows 10 |
| System configuration | CPU: AMD Ryzen 9 5900X |
| | GPU: NVIDIA GeForce RTX 3090 |
| Python library | Cuda 11.7 |
| | Pytorch 1.8 |
| | Torchvision 0.9 |
| | Numpy 1.21 |

### 4.2. Qualitative comparisons with prior works

In Figure 5, we compare the stylized images generated by our method with the stylized images generated by five state-of-the-art style transfer methods. Gatys et al. [4] proposed the seminal optimization-based style transfer method, which generates a stylized image by updating a white noise image. Although some style patterns such as lines and color blocks can be synthesized, many messy artistic effects are generated in the part of the sky in the stylized images (e.g., rows 1, 2, and 6). Similar to our method, Johnson et al. [6] proposed a model-optimization model that is trained to generate images with a single given style image. It can transfer a holistic style texture and structure but sometimes maintains too many local content structures (e.g., rows 1 and 5). AdaIN [8], WCT [9], and SANet [10] are arbitrary style transfer methods that can synthesize stylization results with arbitrary style images. They have a common problem in that they all fail to transfer the essential local style structures such as brushstrokes. Specifically, the color distribution of style images is not maintained accurately in AdaIN (e.g., rows 1, 2, and 4), and too many content structures of stylized images are distorted in WCT (e.g., rows 1, 3, and 8). Compared with AdaIN and WCT, the stylization results generated by SANet have a more appealing visual effect but ignore some details of objects (e.g., rows 4 and 8). In contrast to these methods, our method can simultaneously transfer the local style structure and maintain the color distribution of the style image. Moreover, our stylization results look more natural and cleaner. For example, in the second row, the image of the sky generated by our method looks more artistic and harmonious than other methods in the stylized image.

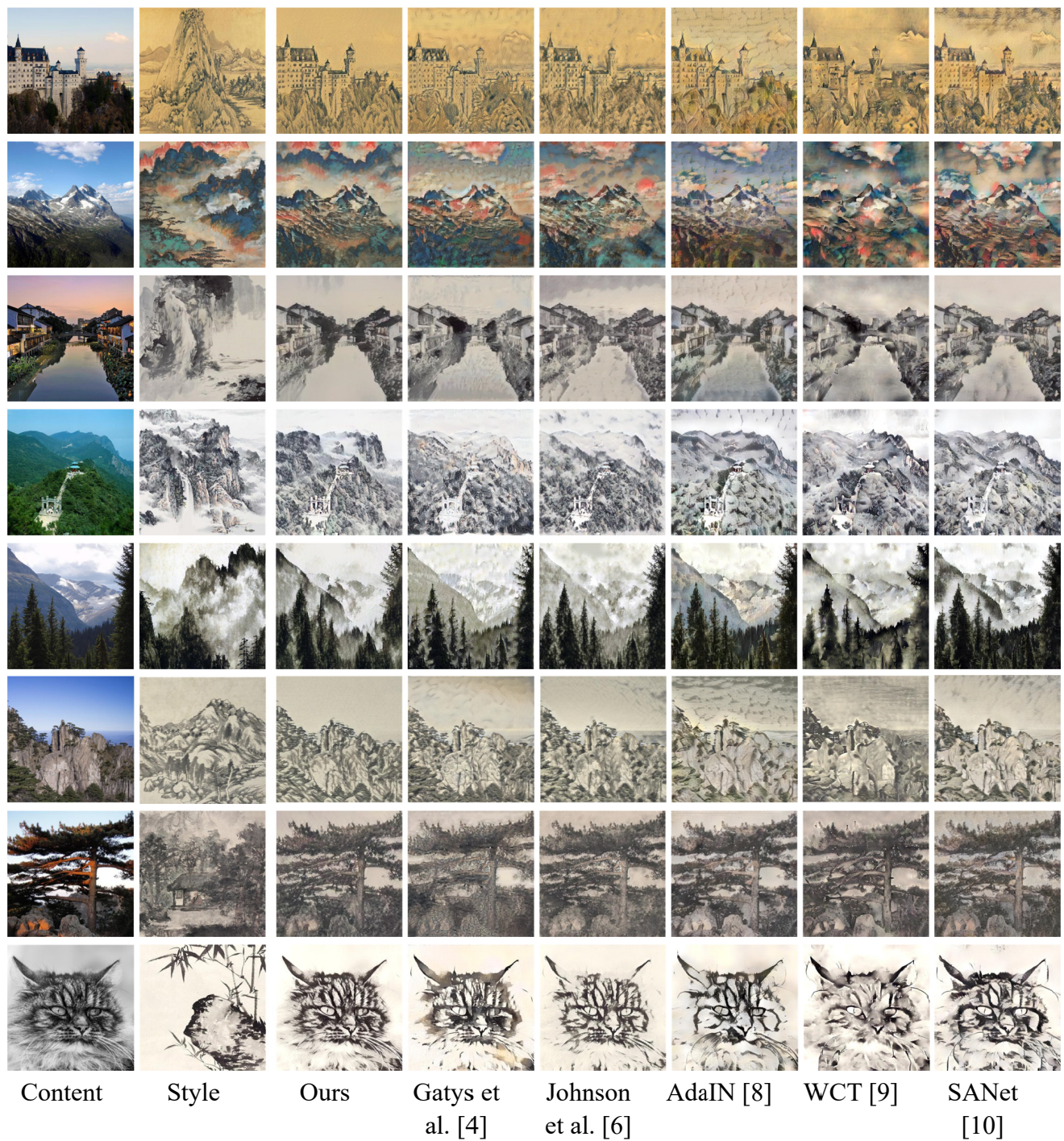

**Figure 5.** Qualitative comparisons between our method and five state-of-the-art methods.

*4.3. Quantitative comparisons with prior works*

In the experiment of quantitative comparisons, we use the peak signal noise ratio (PSNR) to compare the quality of stylized images generated by our method with the quality of stylized images generated by five state-of-the-art methods. A higher PSNR value indicates a higher quality of the reconstructed images. Meanwhile, we use the learned perceptual image patch similarity (LPIPS) proposed in [36] and the structural similarity index measurement (SSIM) proposed in [37] to compute the difference in style similarity between the stylized image and style image. A lower LPIPS value indicates a higher similarity of human perceptual judgments and a higher SSIM value indicates a higher structural style similarity. For each method, 2000 pairs of stylized and style images, which include ten styles, are used to compute the average value of these three metrics respectively.

Table 3 shows that our stylization results have slightly higher image quality than the stylization results generated by other methods when we take PSNR as the metrics. Meanwhile, among these results generated by different methods, our stylization results have the highest style similarity when we use LPIPS and SSIM as the metrics. The experimental results show that our model can synthesize stylized images that have a higher image quality and a higher style similarity to the style images.

**Table 3.** Quantitative comparisons of PSNR, LPIPS, and SSIM between our method and five state-of-the-art methods.

| Methods | Ours | Gatys et al. [4] | Johnson et al. [6] | AdaIN [8] | WCT [9] | SANet [10] |
|---------|------|------------------|--------------------|-----------|---------|------------|
| PSNR | **11.7133** | 11.1994 | 11.5578 | 11.4221 | 11.1369 | 11.2020 |
| LPIPS | **0.6291** | 0.6484 | 0.6533 | 0.6487 | 0.6507 | 0.6440 |
| SSIM | **0.2478** | 0.2261 | 0.2330 | 0.2155 | 0.2144 | 0.2065 |

*4.4. Comparisons of inference speed with prior works*

**Table 4.** Running time comparison between our method and five state-of-the-art methods (in seconds).

| Methods | Ours | Gatys et al. [4] | Johnson et al. [6] | AdaIN [8] | WCT [9] | SANet [10] |
|---------|------|------------------|--------------------|-----------|---------|------------|
| Time | 0.127 | 29.057 | 0.062 | 0.094 | 2.162 | 0.141 |

We compare the inference speed of our proposed method with five state-of-the-art methods. We use different style transfer models to generate 200 stylized images with a resolution of 512 × 512. All experiments are conducted on the same environment configuration. The comparison results are listed in Table 4. Johnson et al. [6] is the faster method because they used only a simple encoder-decoder architecture to achieve style transfer. Our method is the third faster method, and there is only a marginal inference speed difference between our method and the faster method. However, our method can capture multiscale information to synthesize more appealing high-quality stylized images. Compared to other methods except for Johnson et al. [6] and AdaIN [8], our method has the advantage of inference speed. There are two reasons for the fast inference speed: 1) Our Style Transfer Base Network is designed to only synthesize low-resolution coarse stylized images. 2) In the Detail Enhancement Network, only several simple convolution layers and residual blocks are adopted to enhance the details of stylized images. To conclude, our proposed multiscale model traded a small increase in inference time cost for a promising improvement in the quality of stylized images.

## 4.5. Ablation study on loss function

In the first stage, the task of the Style Transfer Base Network is to generate low-resolution coarse stylized images for reconstruction, so we only conduct the ablation experiments to verify the effectiveness of each loss term in the second stage which generates the final stylized images with more details. The results are shown in Figure 6. 1) Without the perceptual loss $l_p$, the basic content structures are discarded such as the shape of the tower disappearing in the stylized image. 2) Without the self-similarity loss $l_{ss}$, some local content details are not transferred and disordered textures are produced; for example, there are some messy brushstrokes in the part of the cloud in the stylized image. 3) Without the mean-variance loss $l_{mv}$, the color distribution is not maintained accurately. 4) Without the rEMD loss $l_r$, key style structures such as some representative brushstrokes are not transferred, and only the color distribution is changed.

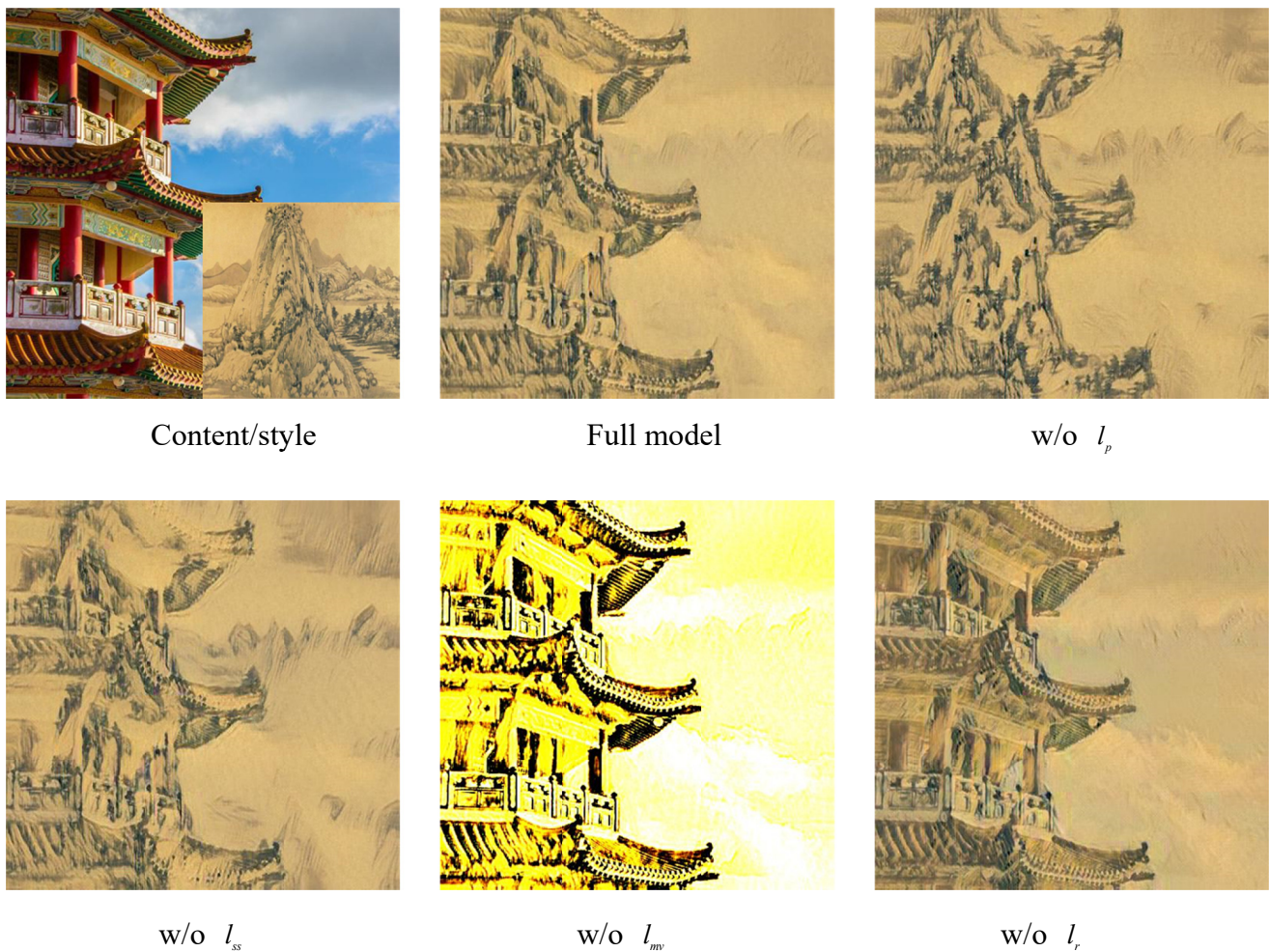

**Figure 6.** Ablation study of effects of different loss terms during training.

*4.6. Effectiveness of Style Transfer Base Network*

We compare the stylized images generated by our full model with the stylized images generated by the model without the Style Transfer Base Network. First, we take $x_c''$ as $x_{cs}''$ when we train the model without the Style Transfer Base Network. In addition, all weight terms are the same for comparing these two models. During of training of two models, Figure 7 shows that the model without the Style Transfer Base Network can achieve a lower content loss after 20,000 iterations because the content image $x_c''$ instead of the stylized image $x_{cs}''$ is fed into the network as input information and used as a low-frequency component for reconstruction. For style loss, the full model can achieve a stable lower value with fewer iterations. As shown in Figure 8, our full model achieves a stylization result that is more in line with the style image. Without the Style Transfer Base Network, the local style structures are not the same as the stylized images. Moreover, the color distribution of stylized images is not transferred because we can only take $x_c''$ as the low-resolution stylized image to reconstruct the final high-resolution stylized image.

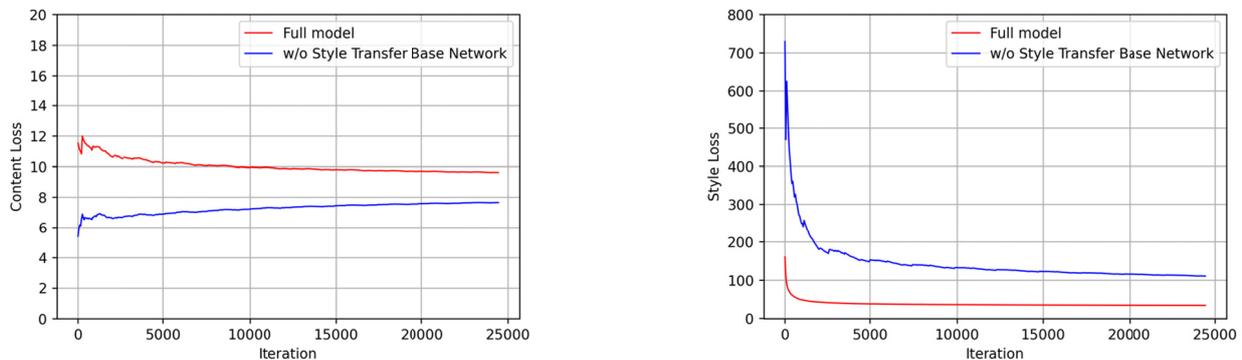

**Figure 7.** Comparison of full model and model without Style Transfer Base Network in terms of content and style losses.

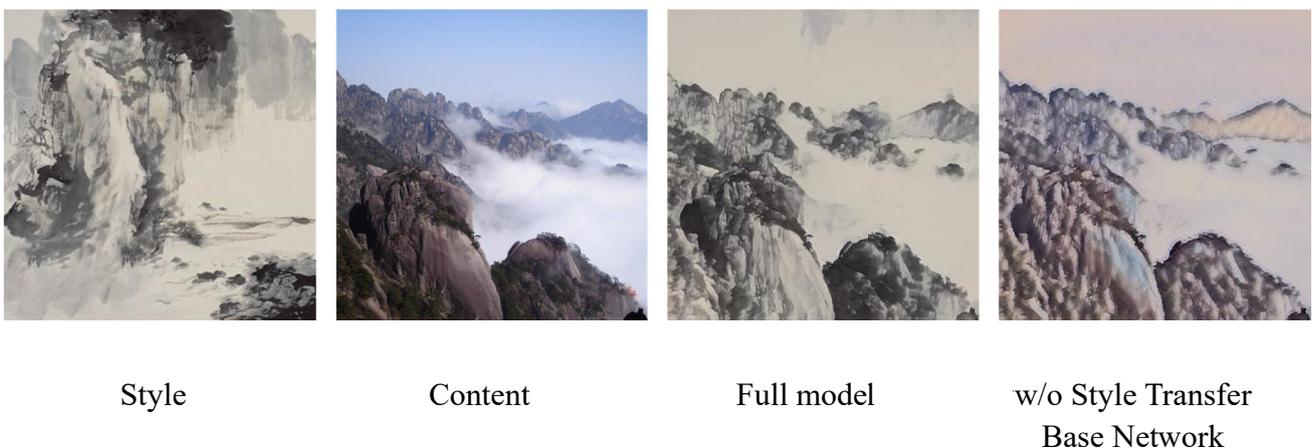

| Style | Content | Full model | w/o Style Transfer Base Network |

**Figure 8.** Comparison of stylized image of full model and the model without Style Transfer Base Network.

*4.7. Effectiveness of Detail Enhancement Network*

We compare the stylized images generated by our full model with the stylized images generated by the model without the Detail Enhancement Network. In Figure 9, we directly take content and style images with a resolution of 512 × 512 as inputs to train the Style Transfer Base Network and then generate unsatisfactory stylized images that lose many local style details. In contrast, our full model with the help of the Detail Enhancement Network can achieve more appealing stylization results that maintain essential style structures and style color distribution.

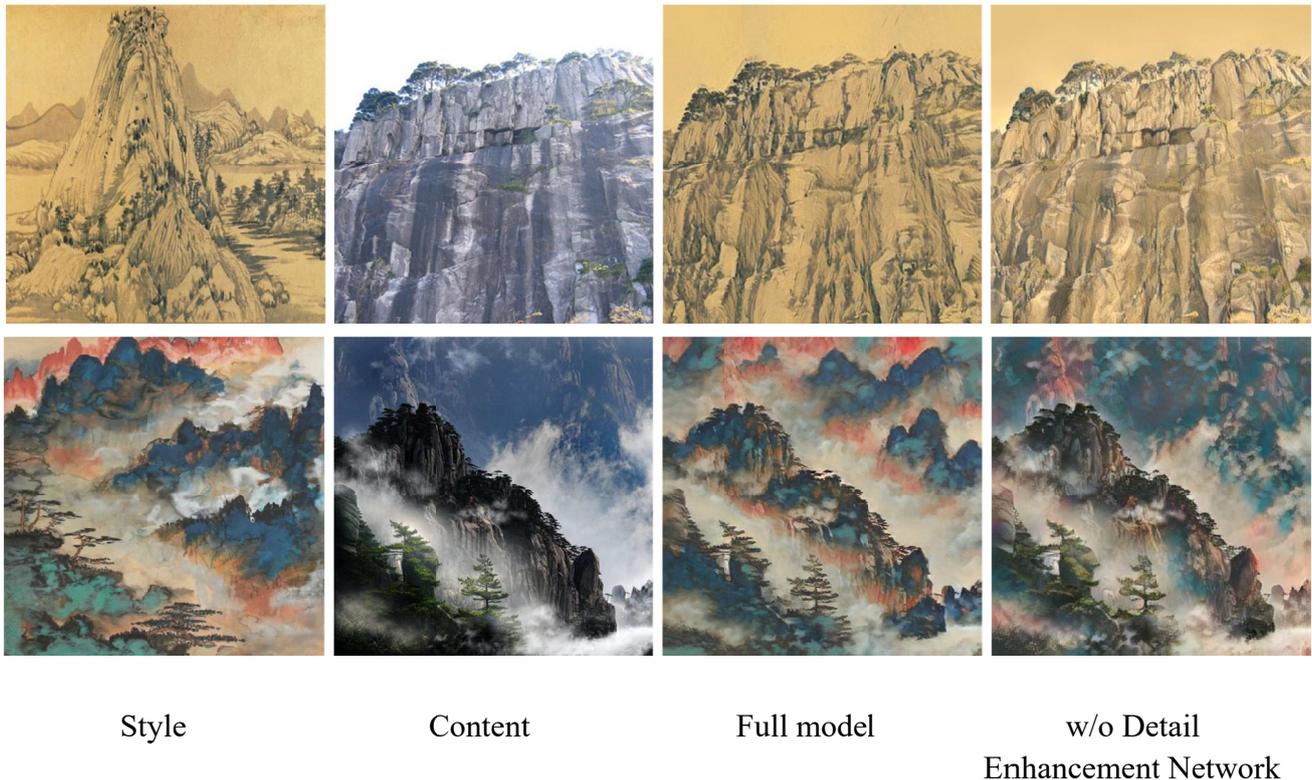

| Style | Content | Full model | w/o Detail Enhancement Network |

**Figure 9.** Comparison of stylized image of full model and the model without Detail Enhancement Network.

*4.8. Effectiveness of EIS module*

We compare the stylized images generated by our full model with the stylized images generated by the model without the EIS module. During the training of the two models, Figure 10 shows that the content and style loss of the model without the EIS module are slightly different from those of the full model between 0 and 5000 iterations. Then, the model without the EIS module obtains a similarity in content and style loss to that of the full model between 5000 and 25,000 iterations of optimization. As shown in Figure 11, our full model with the EIS module can generate more complete stylized images that maintain some details of the content structure. For example, without the EIS module, part of the outline of the mountain disappears in the stylized image in the first row.

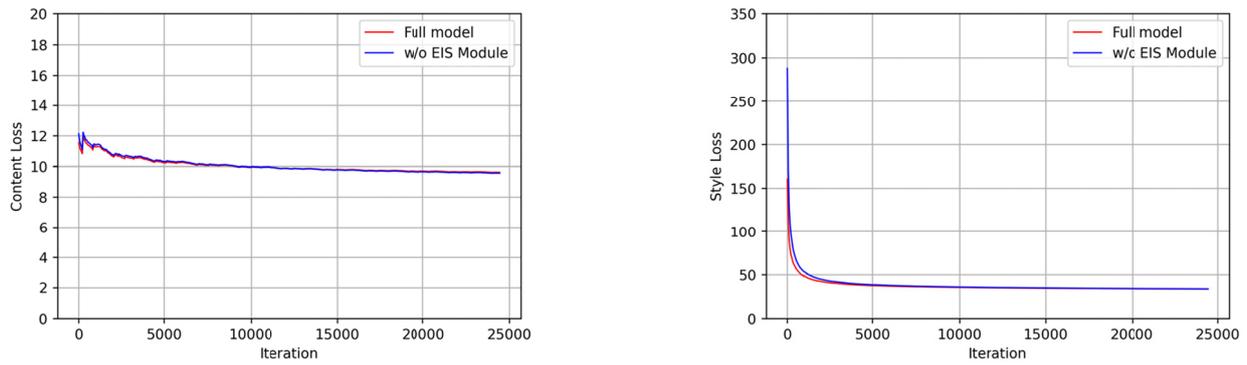

**Figure 10.** Comparison of full model and model without EIS module in terms of content and style losses.

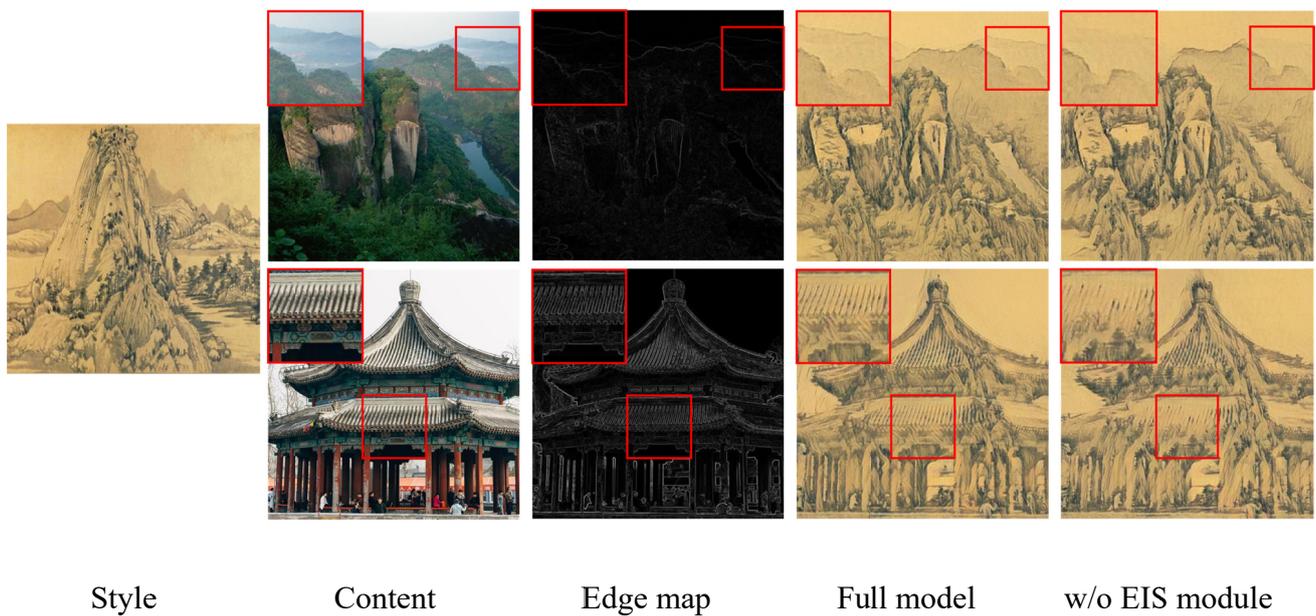

| Style | Content | Edge map | Full model | w/o EIS module |

**Figure 11.** Comparison of full model and the model without EIS module in stylized images.

*4.9. Additional experiments*

To show the high-resolution stylization results generated by our method more clearly, we zoom in on some details in style, content, and stylized images in Figure 12. We can observe that the holistic content structures are transferred to the stylized image while the local texture of our stylized image is extremely similar to that of the style image. For example, the brushstrokes of the vegetation in the stylized image are close to that in the style image (e.g., rows 1 and 3). These results demonstrate that our model can learn the artistic expression of style images.

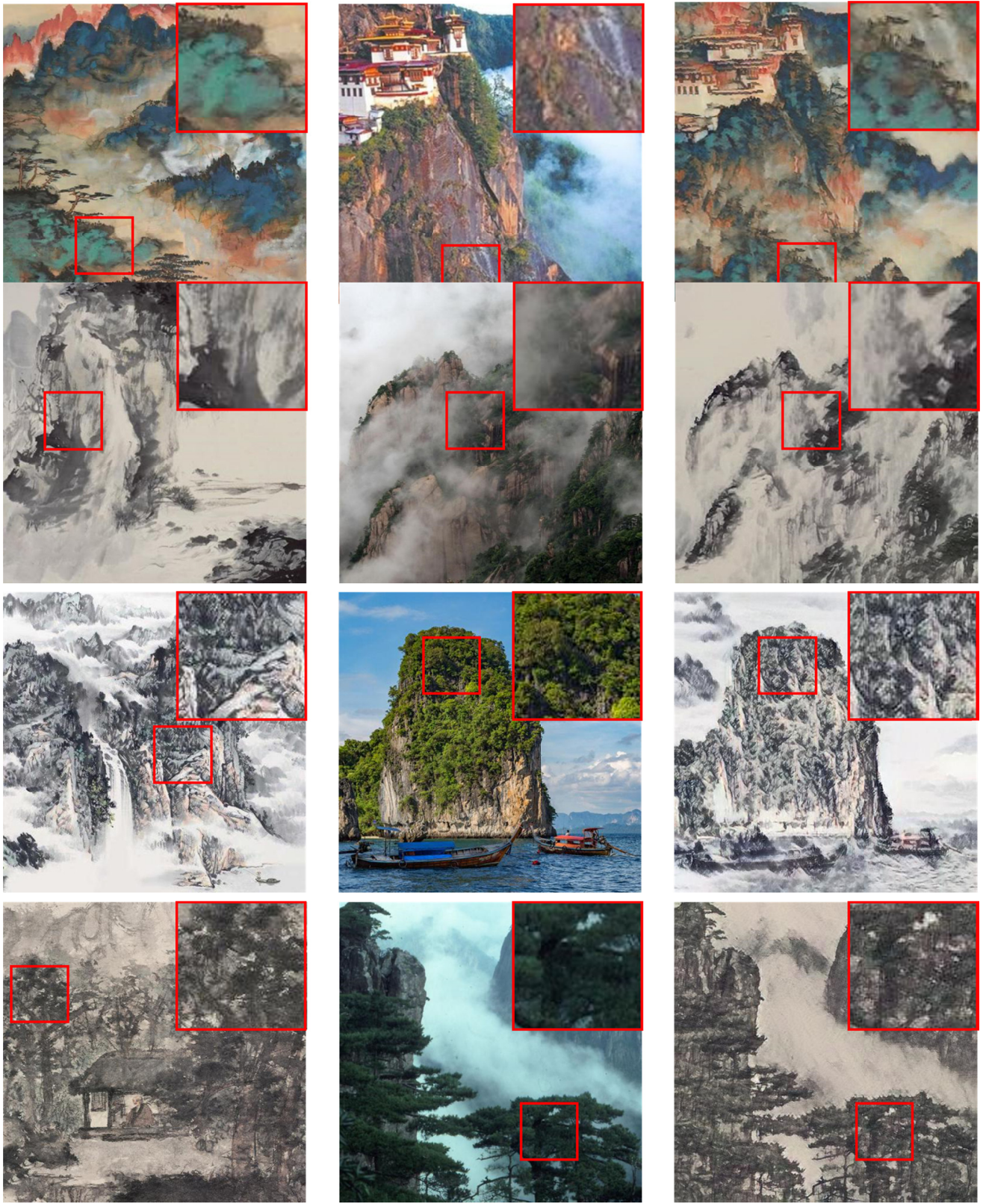

Style                     Content                   Stylization

**Figure 12.** Fine-grained style details of stylization results.

Moreover, in our method, the degree of stylization can be controlled during training by adjusting the weight term $\alpha$. Figure 13 shows that only the color distribution is transferred when we set $\alpha$ to 5 and too many content structures are discarded when we set $\alpha$ to 0.1. The essential style textures are transferred while the basic content structures are maintained when we set $\alpha$ to 1.

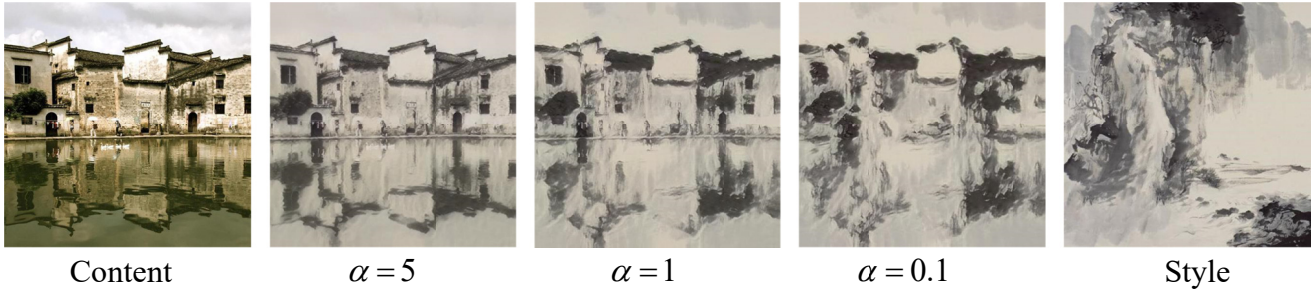

Content $\qquad$ $\alpha = 5$ $\qquad$ $\alpha = 1$ $\qquad$ $\alpha = 0.1$ $\qquad$ Style

**Figure 13.** Trade-off of content-style losses.

## 5.  Conclusions and future work

We proposed a novel efficient style transfer method based on a Laplacian pyramid for Chinese traditional painting. In our model, we adopt Laplacian pyramid decomposition and reconstruction to train our model with multiscale information. First, our Style Transfer Base Network transfers rough style patterns at a low resolution. Then, our Detail Enhancement Network can gradually enhance the details of the content structure and style texture as the resolution increases. In addition, an EIS module is applied in the network to integrate content edge information into our model to select key content structures to improve the visual quality of the stylization results. Eventually, promising high-resolution stylized images can be synthesized by our full model. Different experimental results demonstrated the effectiveness of our full model in synthesizing high-quality stylized images, which are preferred over the stylization results synthesized by other state-of-the-art style transfer methods.

We further discuss the design rationales for our model from three aspects. First, one of the core constructs of our model is Style Transfer Base Network, which can transfer the global coarse style patterns while ignoring the unimportant structural details at a low resolution. Our Style Transfer Base Network is constructed based on the style transfer framework. Therefore, our model can still work when we change our style transfer framework to other style transfer frameworks such as AdaIN or WCT. However, the final generated stylized results are affected by the style transfer framework and different universal style transfer frameworks can be attempted based on different needs. Second, different image-to-image frameworks can be used to replace our simple convolutional networks in Detail Enhancement Network. For example, an encoder-decoder architecture can be applied in our Detail Enhancement Network to generate residual stylized images but it needs more computer computing resources. Finally, we can add an adversarial loss during the optimization of the Detail Enhancement Network. A generative adversarial network objective and a discriminator will consume more computing resources and greatly increase training time, so we do not use them in our model.

Although our method can synthesize high-quality stylization results, only one style of stylized images can be generated after one training process. In future work, we will adopt the framework of our method to achieve the multiscale arbitrary style transfer method which can transfer arbitrary style images to content images after one training process. Moreover, we will try to adopt the adversarial loss to obtain more exquisite and appealing stylized images if our experimental environment gets improved.

## Acknowledgments

This research was funded by the Natural Science Foundation of China (Grant No. 12263008, 62061049), Key R&D Projects in Yunnan Province (Grant No. 202202AD080004), Application and Foundation Project of Yunnan Province (Grant No. 202001BB050032), Department of Science and Technology of Yunnan Province -Yunnan University Joint Special Project for Double-Class Construction (Grant No. 202201BF070001-005), and Postgraduate Practice and Innovation Project of Yunnan University (Grant No. 2021Y177).

## Conflict of interest

The authors declare there is no conflict of interest.